\documentclass[10pt,twocolumn,letterpaper]{article}

\usepackage{wacv}              

\usepackage{xcolor}
\usepackage{booktabs}
\usepackage{amssymb}
\usepackage{algorithm}
\usepackage{algpseudocode}
\usepackage{listings}
\lstdefinestyle{prompt}{%
  basicstyle=\ttfamily\scriptsize,
  breaklines=true,
  breakautoindent=false,
  columns=fullflexible,
  keepspaces=true,
  showstringspaces=false,
  frame=single,
  framesep=3pt,
  xleftmargin=3pt,
  xrightmargin=3pt,
  aboveskip=4pt,
  belowskip=2pt,
}

\usepackage{stfloats}

\definecolor{wacvblue}{rgb}{0.21,0.49,0.74}
\usepackage[pagebackref,breaklinks,colorlinks,allcolors=wacvblue]{hyperref}

\def\wacvPaperID{1062} 
\def\confName{WACV}
\def\confYear{2027}

\title{Open-KNEAD: Knowledge-grounded Nutrition Estimation via Agentic Decomposition}

\author{
Bruce Coburn \qquad Jingbo Yue \qquad Jinge Ma \\ Siddeshwar Raghavan \qquad Gautham Vinod \qquad Fengqing Zhu \\
Purdue University, USA \\
{\tt\small \{coburn6, yue53, ma859, raghav12, gvinod, zhu0\}@purdue.edu}
}

\begin{document}
\maketitle
\begin{abstract}
Multimodal Large Language Models (MLLMs) are increasingly used for dietary assessment from meal images, where retrieval-augmented grounding was shown to sharpen nutrition estimates. However, we find this premise no longer holds for current MLLMs. A modern MLLM's direct estimate now matches or surpasses the full retrieval pipeline. This raises a question: if retrieval no longer improves the overall estimate, can it still deliver the two things clinicians value, accurate portions and a traceable, item-by-item record? We pursue this while preserving what matters for clinical adoption: minimal user burden (a single, unannotated meal image), explainability (an auditable record), and privacy (locally hosted inference). We introduce \textbf{Open-KNEAD}, a knowledge-grounded agentic framework for meal nutrition estimation that is training-free and locally deployable. Each decomposed food item is grounded to a Food and Nutrient Database for Dietary Studies (FNDDS) code via selective, nutrient-aware retrieval, composing an auditable per-item record. Across two open MLLM families and three cuisines, Open-KNEAD improves portion estimates over both prior grounding methods and direct estimation in most backbone-dataset settings. An agent-internal recipe-prior step further recovers the invisible cooking-added energy that biases estimates on non-US cuisine. The advantage is largest on the dietitian-verified ACETADA dataset, where the local open agent surpasses the direct portion estimates of two frontier closed models by $\sim30\%$ and $\sim53\%$, all while keeping every meal image on local hardware. We release the Open-KNEAD framework and its agent-ready FNDDS knowledge base.
\end{abstract}
    
\section{Introduction}
\label{sec:intro}

\begin{figure*}[t]
  \centering
  \includegraphics[width=\textwidth]{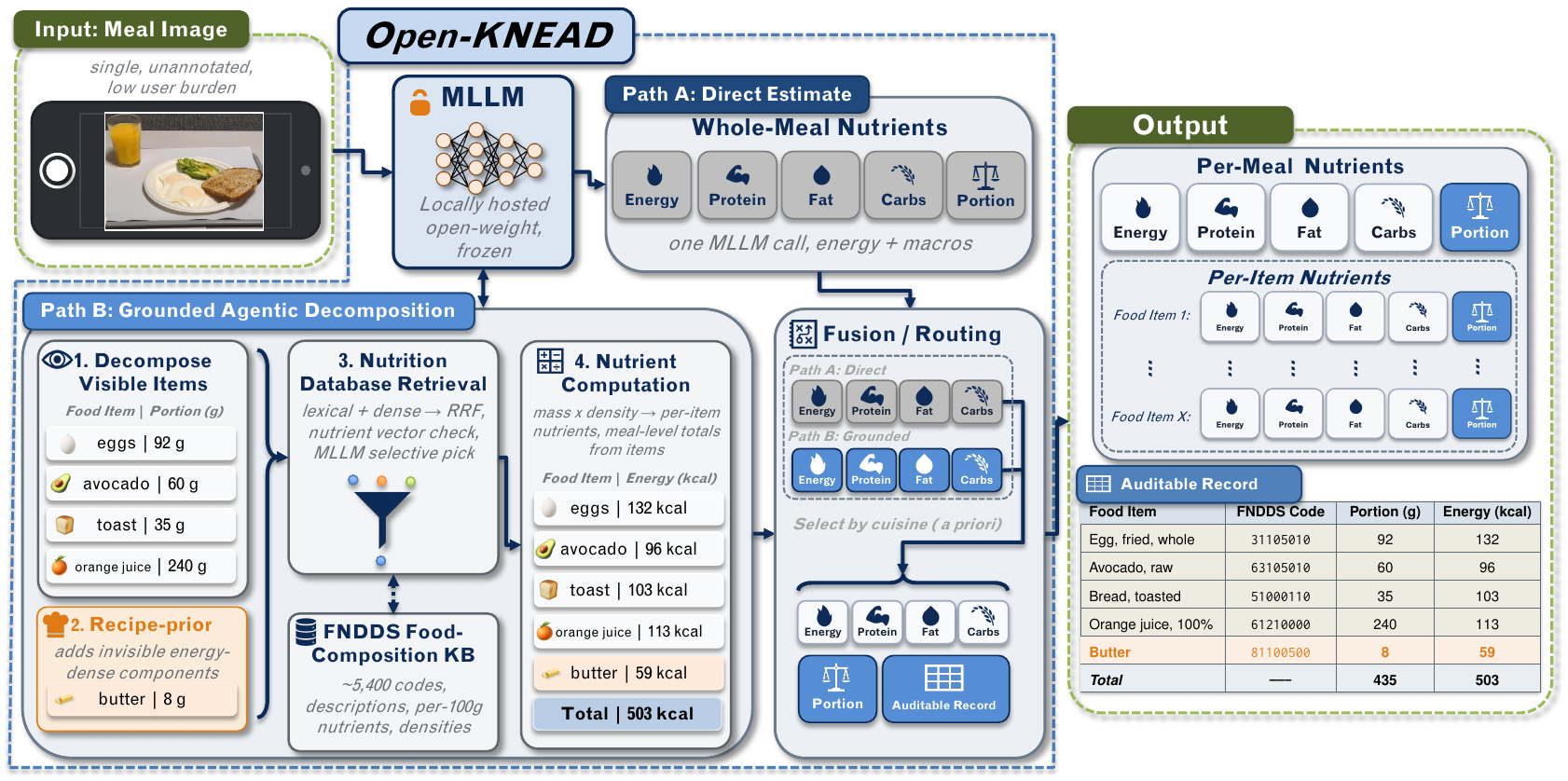}
  \caption{\textbf{Open-KNEAD overview.} Input is a single meal image, processed locally; the output is assembled by routing each quantity to the path that estimates it best.}
  \label{fig:method}
\end{figure*}

Dietary assessment and tracking rely on accurate estimation of the nutritional content of consumed foods and beverages, a task increasingly assigned to Multimodal Large Language Models (MLLMs) leveraging meal images~\cite{romerotapiador2025ready, dietai24, ohara2025chatgpt, tanabe2025reasoning, fridolfsson2025performance, acetada}. While the inherent fine-grained classification is difficult, the primary crux of this task hinges on estimating portion size given a monocular image and accounting for ingredients not visible in the image (i.e., cooking oil or the difference between whole and skim milk).

Retrieval-augmented grounding~\cite{lewis2020rag, hua2024nutribench} once offered a way to sharpen these estimates: Yan et al.~\cite{dietai24} showed that retrieving individual food entries from a nutrition database substantially improved 2023-era estimation. We find this premise no longer holds for current 2026-era MLLMs. Per our evaluation, a modern MLLM's direct estimate (i.e., a single MLLM call that maps image to total macronutrients with no decomposition, retrieval, or knowledge base) either matches or surpasses the full retrieval pipeline. We observe this across retrieval schemes, from naive score-weighted aggregation to learned code selection, as detailed in Section~\ref{sec:results}.

However, grounding provides what a direct estimate cannot: an itemized, database-coded record, and a means to incorporate nutritional knowledge absent from a single image, such as unobserved ingredients. The gap we aim to assess is how to retain these benefits of grounding while counteracting the error that its inclusion introduces.

Beyond accuracy, a deployable system should respect three distinct considerations. First, it should impose minimal user burden~\cite{bailey2021overview, ravelli2020traditional}, requiring only a single, unannotated meal image, with no accompanying labels or metadata, rather than a physical size reference, multiple views, or depth sensors. Next, it should be explainable~\cite{tonekaboni2019clinicians, rudin2019stopexplaining}, producing an itemized, nutrition database-coded record a dietitian can audit rather than a single opaque number. Last, it should preserve privacy~\cite{sharma2020privacy}, using open-weight models that sequester and analyze sensitive images locally rather than sending them through a commercial API. We treat these as firm requirements and ask how a grounded system can meet them without sacrificing accuracy.

We introduce \textbf{Open-KNEAD}, a knowledge-grounded agentic framework that addresses this gap. It decomposes a meal image into its individual food items, mapped to a public, research-quality database, the Food and Nutrient Database for Dietary Studies (FNDDS)~\cite{fndds}, through selective, nutrient-aware retrieval, recovering portion from per-item masses and composing nutrients into an auditable record (\Cref{fig:method}). The agent acts only where it helps: it reserves costly disambiguation for the items whose candidate foods differ enough in nutrient content to change the estimate. To account for ingredients not innately visible in the meal image, such as cooking oil, a leakage-safe recipe-prior infers the hidden ingredients typically contained in a recognized dish and includes them in the estimate. We evaluate across three datasets spanning US, Australian, and Chinese cuisines~\cite{thames2021nutrition5k, acetada, yu2026omnifood8k}. Open-KNEAD is training-free and runs locally on open-weight MLLMs.

We keep Open-KNEAD deliberately minimal to test its inherent capabilities, reserving heavier perception tools and learned portion estimators for future work. Similarly, we do not introduce commercial models in the loop to preserve privacy. In summary, we make three contributions:

\begin{enumerate}
  \item \textbf{Open-KNEAD}, to our knowledge the first knowledge-grounded agentic framework for meal nutrition estimation while being training-free and locally deployable. It improves portion estimation over direct prompting across datasets and open backbones, most strongly on the dietitian-verified ACETADA benchmark, and yields an auditable food item $\to$ food code $\to$ nutrient composition that a direct estimate cannot provide. Running entirely on local hardware, the agent surpasses the direct portion estimates of both frontier closed models (gpt-5.5 and Gemini) on every evaluated dataset, by up to $\sim$53\% on the dietitian-verified benchmark.

  \item \textbf{An agent-ready knowledge base for nutrition estimation, which we release}: we augment FNDDS~\cite{fndds} with dual sparse and dense retrieval embeddings and per-category nutrient-dispersion statistics that drive the agent's selective gating, packaged as a single, public resource for agentic nutrition estimation.

  \item \textbf{The recipe-prior}, a leakage-safe, inference-time method that restores invisible cooking-added ingredients, such as the oil common in Chinese cuisine, reducing systematic underestimation.
\end{enumerate}

\begin{figure*}[htbp]
  \centering
  \includegraphics[width=\textwidth]{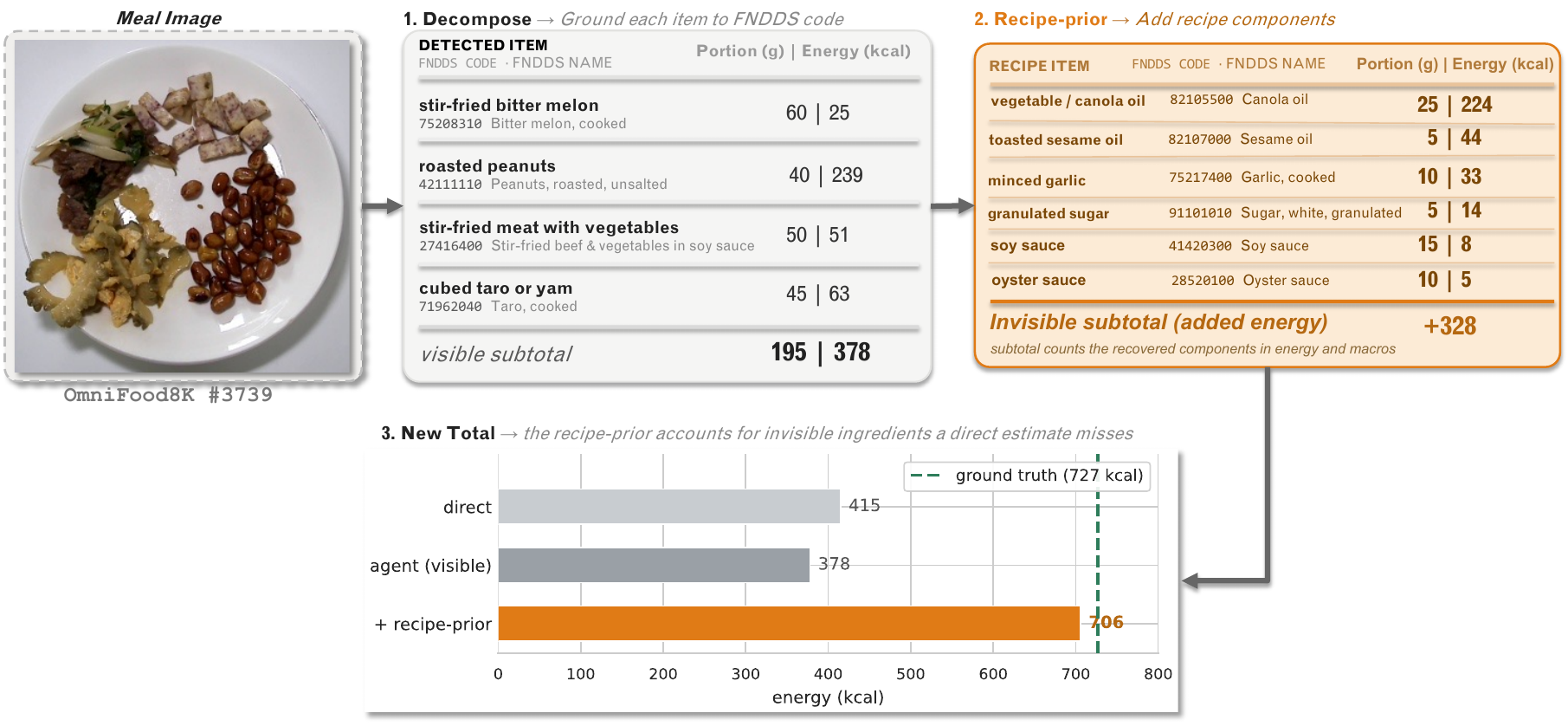}
  \caption{\textbf{Qualitative trace.} One OmniFood meal end to end (Gemma-MoE): the agent grounds each item to an FNDDS code and food name and recovers the invisible cooking oil, lifting energy from 378 to 706 against a 727\,kcal ground truth.}
  \label{fig:trace}
\end{figure*}

\section{Related Work}
\label{sec:related}

Nutrition estimation from meal images has been pursued along three broad directions~\cite{vinod2026pixels}: supervised in-domain regression that maps an image directly to nutrient values~\cite{thames2021nutrition5k, yu2026omnifood8k, qi2025vif2}, geometric portion estimation that estimates or reconstructs food volume~\cite{vinod2024scaling, ma2024mfp3d, chen2026implicit}, and, most recently, MLLM methods that reason over the image through natural-language prompting~\cite{dietai24, romerotapiador2025ready, bhatambarekar2026agentic}. Open-KNEAD belongs to the last of these, advancing it through knowledge grounding and selective agentic reasoning. We review each direction and position our contributions with respect to them.

Within the MLLM line, DietAI24~\cite{dietai24} retrieves food-composition entries to ground a 2023-era model, establishing retrieval augmentation as an effective approach. We reframe and re-implement their method as a controlled comparator, running it on modern MLLMs with the same knowledge base. Retrieval grounding has since been extended across nutrition tasks, over meal descriptions~\cite{hua2024nutribench} and for food identification~\cite{zhou2026nutrirag}. Other work benchmarks modern models for dietary assessment: Romero-Tapiador et al.~\cite{romerotapiador2025ready} at the food-recognition level, and a recent benchmark~\cite{acetada} on energy, macronutrient, and portion estimation that finds contextual metadata such as location and meal time reduces error. We evaluate at the portion, macronutrient, and per-ingredient level, and improve estimation from a single unannotated image without additional context. A recent work by Bhatambarekar and Sarkar~\cite{bhatambarekar2026agentic} prompts a model to self-question agentically, lowering error through interactive user feedback. Open-KNEAD instead grounds each item to a food code from a single unannotated image, disambiguating individual food components without user interaction, and restores invisible ingredients via the recipe-prior.

Supervised regressors map an image directly to nutrients, as in the baselines of Nutrition5k~\cite{thames2021nutrition5k} and OmniFood8K~\cite{yu2026omnifood8k} and visual-ingredient fusion methods such as VIF$^2$~\cite{qi2025vif2}. They are accurate in-distribution but require task-specific training, and their downstream estimates are purely numerical. Open-KNEAD, however, is training-free, applicable across cuisines, and auditable, making it complementary rather than competing.

Geometric methods estimate food volume directly, given either stereo or monocular images. Classical pipelines infer scale from a fiducial marker or a similar reference object within the frame. Recent works, however, relax this requirement through explicit 3D reconstruction: object scaling~\cite{vinod2024scaling}, point-cloud reconstruction~\cite{ma2024mfp3d}, and implicit-scale reconstruction~\cite{chen2026implicit}. Whereas these methods are distinctly geometric, Open-KNEAD grounds portion in nutritional knowledge and is applicable in a monocular, training-free setting. Combining these two paradigms is left for future work.

\section{Method: Open-KNEAD}
\label{sec:method}

We present Open-KNEAD, a training-free framework for grounded meal nutrition estimation. We formalize the task and its design constraints (\S\ref{sec:setting}), describe the nutrition knowledge base that grounds the system (\S\ref{sec:kb}), and detail its two paths, a direct estimate (\S\ref{sec:patha}) and a grounded agentic decomposition (\S\ref{sec:pathb}), combined by routing (\S\ref{sec:fusion}). All prompts are provided in the supplementary material.

\subsection{Problem setting and design constraints}
\label{sec:setting}
Given a single image $x$ of a served meal, Open-KNEAD produces

\begin{equation}
\hat{y} \;=\; \big(\hat{e},\, \hat{y}_{\mathrm{pro}},\, \hat{y}_{\mathrm{fat}},\, \hat{y}_{\mathrm{carb}},\, \hat{m},\, \widehat{\mathcal{C}}\big),
\label{eq:task}
\end{equation}

that is, meal-level energy $\hat{e}$ (kcal), macronutrients $\hat{y}_{\mathrm{pro}}, \hat{y}_{\mathrm{fat}}, \hat{y}_{\mathrm{carb}}$ (g), total mass $\hat{m}$ (g), and an auditable food entry $\widehat{\mathcal{C}} = \{(\text{item}_i, c_i, m_i, \mathbf{n}_i)\}$ mapping each food item to an FNDDS code $c_i$, a mass $m_i$ (g), and a nutrient vector $\mathbf{n}_i$, respectively. To mitigate user burden, the single meal image $x$ is not accompanied by any additional context or modalities (e.g., added views or depth). For privacy, all inference runs locally on an open-weight MLLM $\mathcal{M}$. Open-KNEAD is additionally training-free, using a frozen $\mathcal{M}$ and adding no task-specific fine-tuning.

Open-KNEAD runs two paths from the backbone $\mathcal{M}$. Path A is a direct estimate: a single call returns meal-level energy, macronutrients, and portion. Path B is a grounded agentic decomposition that resolves the meal into individual food items, grounds each to an FNDDS code in the knowledge base, and builds a portion estimate and the auditable food entry $\widehat{\mathcal{C}}$ from per-item masses. Open-KNEAD reports the overall energy and macronutrients together with the decomposition's portion and itemized record. Figure \ref{fig:method} depicts the overall framework, and Algorithm~\ref{alg:Open-KNEAD} details the end-to-end inference procedure.

\subsection{Nutrition Knowledge Base}
\label{sec:kb}
Open-KNEAD grounds items against a frozen food-composition knowledge base of $J\!\approx\!5{,}400$ entries built from the FNDDS \cite{fndds}, the standard reference for US dietary assessment, whose mixed-dish granularity matches how meals are actually logged:

\begin{equation}
\mathcal{K} = \big\{ (c_j,\, d_j,\, \mathbf{n}_j) \big\}_{j=1}^{J},
\label{eq:kb}
\end{equation}

where $c_j$ is the food code, $d_j$ its description, and $\mathbf{n}_j \in \mathbb{R}^{4}$ the per-100\,g nutrient vector (energy, protein, fat, carbohydrate).

Each entry also carries a What We Eat in America (WWEIA) food category $g(j)$~\cite{wweia}, the USDA scheme that groups foods into mutually exclusive categories of similar type and nutrient content (e.g., one category for cheeses, another for fried chicken). Over each category we precompute a per-nutrient dispersion vector $\boldsymbol{\sigma}_g \in \mathbb{R}^4$, the within-category standard deviation of $\mathbf{n}$. Normalizing in this way allows the agent to gate on nutritional ambiguity rather than raw magnitude. Essentially, the macronutrients' dispersions are measured within a category rather than globally, registering the differences that matter for a given food (a few tens of kcal per 100\,g, trivial across the database, can separate two cuts of meat). We floor $\boldsymbol{\sigma}_g$ at a fraction of the global dispersion so that categories of near-identical foods, whose internal dispersion is otherwise near zero, do not register as spuriously ambiguous.

We further index $\mathcal{K}$ twice, a sparse lexical index over the descriptions (leveraging BM25 \cite{robertson2009bm25}) and a dense index of their embeddings \cite{karpukhin2020dpr}, using a swappable open-weights encoder to preserve the local-inference setting. We release this packaged resource (records, dual indices, dispersion statistics, portion and density priors) as the \textbf{agent-ready knowledge base}. The framework is agnostic to this choice of database, so other food-composition databases can be substituted in the same pipeline.

\subsection{Path A: Direct Estimate}
\label{sec:patha}
A single call $\mathcal{M}(x)$ yields the direct estimate $\hat{y}^{A} = (\hat{e}^{A}, \hat{y}^{A}_{\mathrm{pro}}, \hat{y}^{A}_{\mathrm{fat}}, \hat{y}^{A}_{\mathrm{carb}}, \hat{m}^{A})$, estimating energy, macronutrients, and total portion jointly from the image. Path A provides Open-KNEAD's base energy and macronutrient estimate and doubles as the standalone direct baseline we evaluate against. It also predicts a portion $\hat{m}^{A}$, but Open-KNEAD instead takes the portion from Path B's decomposition, which estimates it more accurately (\S\ref{sec:r1}).

\subsection{Path B: Grounded Agentic Decomposition}
\label{sec:pathb}
Path B is primarily responsible for the portion estimate and the itemized composition. It begins by decomposing the image into visible food components, a single call returning $\mathcal{I}_{\mathrm{vis}} = \{(\text{name}_i, m_i)\}_{i=1}^{k} \leftarrow \mathcal{M}(x)$, each visible food item with an estimated portion in grams. We ask only for visible items, and place no cap on how many the model may list, to mirror the granularity of dietary logging.

However, this visible-only decomposition can miss energy-dense components an image does not show, chiefly added fats and sugars such as cooking oil, butter, sauces, and added sugar. We recover them with a recipe-prior, an additional text-only call to $\mathcal{M}$ that returns the cooking-added ingredients a dish of the recognized items typically contains:

\begin{equation}
\mathcal{I} \;=\; \mathcal{I}_{\mathrm{vis}} \,\cup\, \Phi(\mathcal{I}_{\mathrm{vis}}) \cdot \mathcal{R}\!\big(\mathrm{names}(\mathcal{I}_{\mathrm{vis}})\big),
\label{eq:recipe}
\end{equation}

where $\mathcal{R}(\cdot)$ supplies these \emph{typical} invisible additions with approximate masses, and the binary gate $\Phi \in \{0,1\}$ turns the step off ($\Phi = 0$) for meals with no cooking-added components, such as a piece of fruit or an undressed salad.

This remains a prior, rather than a ground-truth lookup, for two reasons. It is leakage-safe, conditioned only on the recognized item names, never the meal's true recipe and by default not the image; and composite-gated, returning nothing for a plate of fruit or an undressed salad, so it cannot add oil to an apple. Recovered items, for example cooking oil and butter for a meal of fried chicken, white rice, and saut\'eed green beans, are appended and grounded to FNDDS codes like any visible item, contributing their full per-100\,g nutrient vector through Eq.~\ref{eq:aggregate} rather than energy alone (for oil and butter, chiefly fat).

Each item, visible or recipe-added, queries both indices, which are complementary: the lexical retriever matches exact food-name terms, while the dense retriever matches semantically, retrieving entries whose wording differs from the query. We fuse their rankings by reciprocal rank fusion (RRF \cite{cormack2009rrf}), summing each candidate's reciprocal rank in the two lists so a food ranked highly by either list is kept, with no need to calibrate the two score scales:

\begin{equation}
s_i(c) \;=\; \sum_{r \,\in\, \{\mathrm{lex},\, \mathrm{dense}\}} \frac{1}{\kappa + \mathrm{rank}_r(c)}, \qquad \kappa = 60,
\label{eq:rrf}
\end{equation}

which gives a top-$K$ candidate set $\mathcal{C}_i$ with weights $w_c \propto s_i(c)$.

Rather than call $\mathcal{M}$ on every item, Open-KNEAD first scores how much the candidates could disagree on the nutrients that matter, a retrieval-weighted, dispersion-normalized variance:

\begin{equation}
V_i \;=\; \sum_{c \in \mathcal{C}_i} w_c \, \big\Vert (\mathbf{n}_c - \bar{\mathbf{n}}_i) \oslash \boldsymbol{\sigma}_{g(i)} \big\Vert_2^2, \qquad \bar{\mathbf{n}}_i = \textstyle\sum_{c} w_c \mathbf{n}_c .
\label{eq:variance}
\end{equation}

If $V_i < \tau$ the candidates are nutritionally interchangeable and the weighted mean $\bar{\mathbf{n}}_i$ commits with no further call; only when $V_i \geq \tau$ does $\mathcal{M}$ pick $c_i \in \mathcal{C}_i$, under a bounded per-meal budget $B$ ($B = 5$). This spends the model's calls only where the choice is nutritionally consequential.

Per-item nutrients then scale by mass and sum:

\begin{equation}
\hat{\mathbf{y}}^{B} = \sum_{i \in \mathcal{I}} \frac{m_i}{100}\, \mathbf{n}_{c_i}, \qquad \hat{m} = \sum_{i \in \mathcal{I}_{\mathrm{vis}}} m_i ,
\label{eq:aggregate}
\end{equation}

where $\hat{m}$ sums \emph{visible} items only, since recipe-added components add nutrients but no separately measurable mass. Path B yields a portion estimate and an item $\to$ code $\to$ mass $\to$ nutrient record auditable in a way the direct estimate is not.

\subsection{Routing the two paths}
\label{sec:fusion}
Open-KNEAD assembles its final output from the two paths:

\begin{equation}
\underbrace{\hat{e},\, \hat{y}_{\mathrm{pro}},\, \hat{y}_{\mathrm{fat}},\, \hat{y}_{\mathrm{carb}}}_{\text{Path A or Path B}+\text{recipe (routed)}}, \qquad
\underbrace{\hat{m}}_{\text{Path B (visible)}}, \qquad
\underbrace{\widehat{\mathcal{C}}}_{\text{Path B}} .
\label{eq:fusion}
\end{equation}

The portion and the auditable food entry $\widehat{\mathcal{C}}$ always come from Path B's decomposition. Energy and macronutrients are routed by how well the knowledge base covers the cuisine. We fix this rule a priori from the knowledge base rather than tuning it on test labels: because FNDDS catalogs US foods, its per-100\,g densities transfer to in-distribution US meals but carry a systematic density bias on cuisines it covers poorly, which inflates the decompose-sum total on the off-distribution Australian dataset. Open-KNEAD reports Path B's grounded energy where coverage is good, in-distribution Nutrition5k and OmniFood once the recipe-prior restores its invisible energy, and falls back to Path A's lower-variance direct estimate where it is not, off-distribution ACETADA (\S\ref{sec:r1}). The recipe-prior's recovered components enter the Path B totals only on recognized cuisines where such invisible additions are typical (the gate $\Phi$ of Eq.~\ref{eq:recipe}); a leakage-safe per-meal gate does not improve on this cuisine-level rule.

\begin{algorithm}[t]
\caption{Open-KNEAD inference (one meal, training-free)}
\label{alg:Open-KNEAD}
\begin{algorithmic}[1]
\Require image $x$; frozen MLLM $\mathcal{M}$; KB $\mathcal{K}$; budget $B$; gate threshold $\tau$
\State $\hat{y}^{A} \gets \mathcal{M}(x)$ \Comment{Path A: Direct Estimate}
\State $\mathcal{I}_{\mathrm{vis}} \gets \mathcal{M}(x)$ \Comment{visible items + grams}
\State $\mathcal{I} \gets \mathcal{I}_{\mathrm{vis}} \cup \Phi\!\cdot\!\mathcal{R}(\mathrm{names})$ \Comment{recipe-prior, Eq.~\ref{eq:recipe}}
\For{item $i \in \mathcal{I}$}
  \State $\mathcal{C}_i \gets$ top-$K$ by RRF over BM25 + dense \Comment{Eq.~\ref{eq:rrf}}
  \State compute ambiguity $V_i$ \Comment{Eq.~\ref{eq:variance}}
  \If{$V_i < \tau$ \textbf{or} budget $B$ exhausted}
     \State commit weighted $\bar{\mathbf{n}}_i$
  \Else
     \State $c_i \gets \mathcal{M}(x, \mathcal{C}_i)$; \; $B \gets B-1$ \Comment{selective pick}
  \EndIf
\EndFor
\State $\hat{\mathbf{y}}^{B}, \hat{m} \gets$ aggregate \Comment{Eq.~\ref{eq:aggregate}}
\State \Return route$(\hat{y}^{A}, \hat{\mathbf{y}}^{B}, \hat{m},
\widehat{\mathcal{C}})$ \Comment{Eq.~\ref{eq:fusion}}
\end{algorithmic}
\end{algorithm}

\begin{table*}[htbp]
\centering
\caption{\textbf{Main comparison} (Gemma-MoE; full test splits). Energy MAE, pMAE, and severe-error rate (sev); portion MAE and pMAE. \textbf{Bold} = best, \underline{underline} = second best per dataset; $\downarrow$ lower is better.}
\label{tab:main}
\small
\setlength{\tabcolsep}{6pt}
\begin{tabular*}{\linewidth}{@{\extracolsep{\fill}}l ccc cc}
\toprule
& \multicolumn{3}{c}{\textbf{Energy}} & \multicolumn{2}{c}{\textbf{Portion}} \\
\cmidrule(lr){2-4}\cmidrule(lr){5-6}
Method & MAE ($\downarrow$) & pMAE\% ($\downarrow$) & sev\% ($\downarrow$) & MAE ($\downarrow$) & pMAE\% ($\downarrow$) \\
\midrule
\multicolumn{6}{l}{\emph{Nutrition5k (US)}} \\
Direct prompting                            & 121.9 & 48.1 & 38.8 & 80.2 & 41.8 \\
Self-consistency direct~\cite{wang2023selfconsistency} & 121.6 & 48.0 & 39.1 & 80.1 & 41.7 \\
Structured decomposition (no KB)            & 111.6 & 44.1 & 31.1 & 71.2 & 37.1 \\
RAG, weighted top-$K$~\cite{lewis2020rag}   & 126.3 & 49.8 & 38.0 & 67.7 & 36.3 \\
RAG\,$+$\,VLM pick~\cite{dietai24}          & \underline{111.2} & \underline{43.9} & \underline{28.8} & \underline{66.9} & \underline{35.8} \\
DietAI24~\cite{dietai24}$^{\dagger}$        & 164.9 & 52.9 & 54.6 & 123.7 & 46.7 \\
\textbf{Open-KNEAD (ours)}$^{\S}$           & \textbf{109.7} & \textbf{43.3} & \textbf{28.7} & \textbf{66.7} & \textbf{34.7} \\
\midrule
\multicolumn{6}{l}{\emph{OmniFood8K (CN)}} \\
Direct prompting                            & \underline{155.2} & \underline{40.6} & \underline{61.2} & 62.4 & 30.2 \\
Self-consistency direct                     & 155.4 & 40.7 & 61.7 & 61.9 & 29.9 \\
Structured decomposition (no KB)            & 165.2 & 43.2 & 62.5 & 45.5 & \underline{22.0} \\
RAG, weighted top-$K$                        & 181.6 & 47.5 & 66.7 & \underline{44.6} & 22.3 \\
RAG\,$+$\,VLM pick                          & 184.1 & 48.2 & 66.9 & 45.7 & 23.0 \\
DietAI24$^{\dagger}$                        & 224.6 & 57.9 & 76.0 & 102.0 & 50.7 \\
\textbf{Open-KNEAD (ours)}$^{\S}$           & \textbf{139.1} & \textbf{36.4} & \textbf{50.7} & \textbf{43.5} & \textbf{21.0} \\
\midrule
\multicolumn{6}{l}{\emph{ACETADA (AU)}} \\
Direct prompting                            & \underline{163.5} & \underline{24.2} & \underline{59.6} & 254.1 & 30.5 \\
Self-consistency direct                     & 163.8 & 24.6 & 60.0 & 255.1 & 30.6 \\
Structured decomposition (no KB)            & \textbf{162.9} & \textbf{24.1} & \textbf{56.3} & \underline{174.4} & 20.9 \\
RAG, weighted top-$K$                        & 222.7 & 32.9 & 72.7 & 175.4 & 20.9 \\
RAG\,$+$\,VLM pick                          & 186.9 & 27.6 & 63.5 & 174.6 & \underline{20.8} \\
DietAI24$^{\dagger}$                        & 250.6 & 37.5 & 71.9 & 368.7 & 45.6 \\
\textbf{Open-KNEAD (ours)}$^{\S}$           & \underline{163.5} & \underline{24.2} & \underline{59.6} & \textbf{167.3} & \textbf{20.1} \\
\bottomrule
\end{tabular*}
\\[2pt]
{\footnotesize $^{\S}$ Routed: portion from the decomposition; energy by cuisine (\S\ref{sec:fusion}). \quad $^{\dagger}$ Independent faithful reimplementation; own run and $n$.}
\end{table*}

\section{Results}
\label{sec:results}

After detailing the evaluation setup (\S\ref{sec:setup}), we report three findings: the grounded agent sharpens portion estimates, surpassing even frontier closed models while keeping inference local (\S\ref{sec:r1}); an agent-internal recipe-prior recovers the invisible cooking-added energy that biases non-US cuisine (\S\ref{sec:r3}); and grounding yields an auditable per-ingredient record at lower per-item error (\S\ref{sec:r4}). Ablations isolate the contribution of each component (\S\ref{sec:ablations}).

\subsection{Experimental Setup}
\label{sec:setup}

\noindent\textbf{Datasets (three cuisines).} We evaluate on three single-image meal datasets spanning distinct cuisines: Nutrition5k (US; we additionally report an N5k$^{*}$ subset of $\geq$2-ingredient meals that aligns with DietAI24~\cite{dietai24}), ACETADA (Australian; dietitian-verified served totals), and OmniFood8K (Chinese). We use the full test splits with a fixed view-selection seed. Table~\ref{tab:datasets} situates them: most food datasets target recognition or segmentation and lack the nutrient, mass, and per-ingredient ground truth an agentic decomposition needs; only these three provide it on real plated meals.

\begin{table}[t]
\centering
\caption{\textbf{The food-dataset landscape.} Agentic decomposition needs real plated meals with measured nutrients, mass, \emph{and} per-ingredient ground truth; only the three we adopt (bottom) provide all four, while representative others each lack at least one (full list in the supplementary material). ($\sim$ = partial.)}
\label{tab:datasets}
\small
\setlength{\tabcolsep}{4pt}
\begin{tabular}{l c c c c}
\toprule
Dataset & \shortstack{Plated\\meals} & \shortstack{Measured\\nutrients} & \shortstack{Portion/\\mass GT} & \shortstack{Per-ingred.\\GT} \\
\midrule
Food-101~\cite{food101}            & $\times$    & $\times$ & $\times$ & $\times$ \\
FoodSeg103~\cite{foodseg103}       & \checkmark  & $\times$ & $\times$ & $\sim$ \\
ECUSTFD~\cite{ecustfd}             & $\times$    & $\times$ & \checkmark & $\times$ \\
MetaFood3D~\cite{metafood3d}       & $\times$    & \checkmark & \checkmark & $\sim$ \\
\midrule
\textbf{Nutrition5k}~\cite{thames2021nutrition5k} & \checkmark & \checkmark & \checkmark & \checkmark \\
\textbf{ACETADA}~\cite{acetada} & \checkmark & \checkmark & \checkmark & \checkmark \\
\textbf{OmniFood8K}~\cite{yu2026omnifood8k} & \checkmark & \checkmark & \checkmark & \checkmark \\
\bottomrule
\end{tabular}
\end{table}

\noindent\textbf{Backbones.} We evaluate two open-weight families at multiple scales: Gemma~\cite{gemma} (E4B-it, 31B-it, and a 26B-A4B-it MoE) and Qwen~\cite{qwen} (9B and a 35B-A3B MoE). Two frontier closed models, gpt-5.5 and Gemini, serve only as references for direct estimation; we anchor the comparison to gpt-5.5, the GPT lineage that prior retrieval-based work~\cite{dietai24} was built on.

\noindent\textbf{Comparators.} We compare against direct prompting; self-consistency direct ($K{=}5$), the cost-matched control; structured (ungrounded) decomposition; naive RAG aggregation; and RAG with VLM pick, the DietAI24~\cite{dietai24} analog on the same models and knowledge base. \textbf{Open-KNEAD} is the full routed system, whose internal paths appear as ablations in Table~\ref{tab:ablations}. A faithful replication of the DietAI24 pipeline validates our reimplementation. Supervised in-domain regressors~\cite{thames2021nutrition5k,yu2026omnifood8k,qi2025vif2} train on the test distribution and yield no auditable composition, a different regime; we cite but do not run them.

\noindent\textbf{Implementation.} The open-weight backbones run locally with vLLM, each on a single NVIDIA H100 (Slurm-scheduled), and no meal image leaves the host, satisfying the privacy constraint. Retrieval fuses a sparse BM25 index and a dense index by reciprocal rank fusion ($\kappa{=}60$) and passes the top $K{=}10$ candidates per item to the selective gate; the dense encoder is the open-weight BGE-M3 model~\cite{bge_m3}, run on the same local host. The gate threshold $\tau{=}0.5$ and the per-meal pick budget $B{=}5$ are held fixed across all datasets and backbones, with no per-dataset tuning. Inference is non-thinking and resume-safe; the closed models (gpt-5.5, Gemini) are queried only for direct estimation.

\begin{figure*}[htbp]
  \centering
  \includegraphics[width=\textwidth]{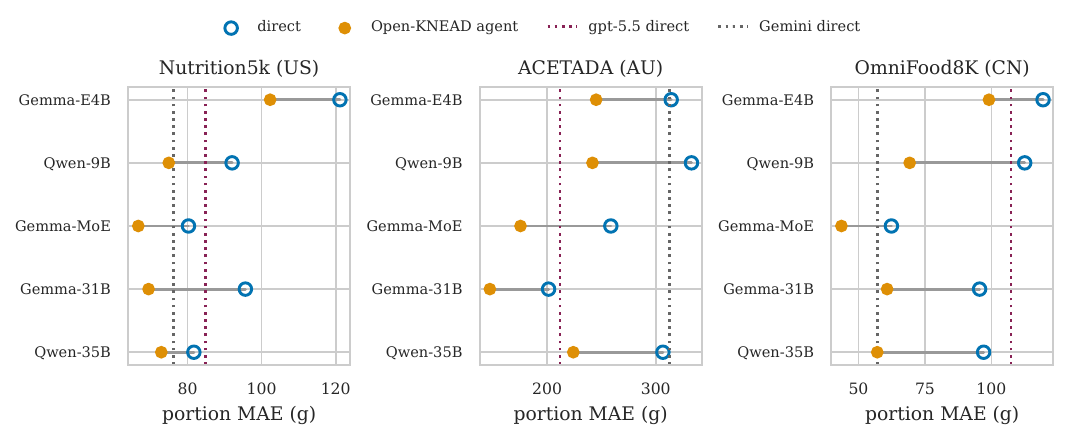}
  \caption{\textbf{Portion MAE, direct vs.\ Open-KNEAD agent, across every backbone $\times$ dataset.} Each backbone is a direct estimate (dot) connected to the agent estimate. Dotted lines mark frontier closed-model direct estimates (gpt-5.5, Gemini).}
  \label{fig:portionbars}
\end{figure*}

\noindent\textbf{Metrics.} For each nutrient $y \in \{$kcal, portion, protein, fat, carb$\}$ over $N$ meals:
\begin{equation}
\mathrm{MAE} = \tfrac{1}{N}\textstyle\sum_n |\hat{y}_n - y_n|, \qquad
\mathrm{pMAE} = 100 \cdot \tfrac{\mathrm{MAE}}{\bar{y}},
\label{eq:metrics}
\end{equation}
plus, for the cross-cuisine analysis, the \emph{signed bias} $\tfrac{1}{N}\sum_n (\hat{y}_n - y_n)$. We use pMAE (MAE over the mean ground truth, the Nutrition5k/DietAI24 convention~\cite{thames2021nutrition5k,dietai24}) rather than per-sample mean absolute percentage error (MAPE), which explodes on near-zero macronutrient ground truths. All metrics are computed at meal \emph{and} ingredient level, and every reported difference carries a paired bootstrap 95\% CI~\cite{efron1993bootstrap}.

Table~\ref{tab:main} compares methods on a representative backbone across energy accuracy, energy \emph{consistency} (the severe-error rate, the fraction of meals off by more than 100\,kcal), and portion. Direct and self-consistency are weak on portion; the retrieval baselines, including the faithful DietAI24 reimplementation, on energy; and the ungrounded decomposition, though competitive on ACETADA energy, returns no provenance. Open-KNEAD leads Nutrition5k and OmniFood on both energy accuracy and consistency and is best-tier on portion throughout, while uniquely returning an auditable per-ingredient composition; only on ACETADA energy does it trail the ungrounded decomposition.

\subsection{Portion estimation and the privacy result}
\label{sec:r1}

The grounded agent improves portion over the direct estimate in 12 of 15 backbone-dataset settings (all CIs exclude zero), with the largest gains on dietitian-verified ACETADA (Figure~\ref{fig:portionbars}; per-backbone numbers in the supplementary material). The three exceptions are near-ties whose CIs span zero: Gemma-E4B on Nutrition5k and OmniFood (decomposition too coarse to refine the direct portion), and Gemma-31B on ACETADA (direct portion already strong). Portion benefits because it is recovered additively from per-item masses, which the decomposition estimates well, whereas energy and macronutrients multiply each mass by a grounded per-100\,g density, so density error and the knowledge base's US-centric bias compound on top of the mass error.

The advantage holds against frontier closed models. On the dietitian-verified set, the strongest open model's local agent (Gemma-31B) reaches 147\,g portion MAE, beating gpt-5.5's direct estimate (212\,g, $-$30\%) and Gemini's (311\,g, $-$53\%) while keeping every image on the local host (Figure~\ref{fig:portionbars}).

\subsection{Cross-cuisine energy: the recipe-prior}
\label{sec:r3}

\begin{table}[htbp]
\centering
\caption{\textbf{The recipe-prior helps only where energy is hidden} (energy MAE; Gemma-MoE; agent = raw decompose-sum, pre-routing). \textbf{Bold} = the energy source Open-KNEAD deploys (routed by cuisine).}
\label{tab:recipe}
\small
\begin{tabular}{l ccc}
\toprule
& \multicolumn{3}{c}{Energy MAE (kcal)\,$\downarrow$} \\
\cmidrule(lr){2-4}
Dataset & direct & agent & agent\,$+$\,recipe \\
\midrule
OmniFood8K (CN)  & 155.2 & 168.6 & \textbf{139.1} \\
Nutrition5k (US) & 121.9 & \textbf{109.7} & 209.7 \\
ACETADA (AU)     & \textbf{163.5} & 198.5 & 309.4 \\
\bottomrule
\end{tabular}
\end{table}

Many cuisines add energy-dense ingredients during cooking, such as oil, butter, or sugar, that a single image never shows, so a visible-only decomposition underestimates their energy. The recipe-prior recovers these for any such meal, reintroducing the typical invisible additions from the recognized items alone and grounding them like a visible item. The effect is clearest where this hidden energy is both substantial and unmeasured, across our benchmarks the Chinese dishes of OmniFood, where every backbone underestimates energy by 50 to 130\,kcal; restoring it lowers energy MAE to 139.1\,kcal on the deployed Gemma-MoE backbone, below the direct estimate's 155.2 (Table~\ref{tab:recipe}). Because the prior only adds energy, applying it to a meal with nothing hidden would inflate the estimate, so it acts selectively: the per-meal gate $\Phi$ turns it off for meals with no cooking-added components, and cuisine-level routing adds its recovered energy to the deployed total only where such additions are typical (Table~\ref{tab:recipe}).

\subsection{Interpretable per-ingredient grounding}
\label{sec:r4}

\begin{figure}[htbp]
  \centering
  \includegraphics[width=0.8\linewidth]{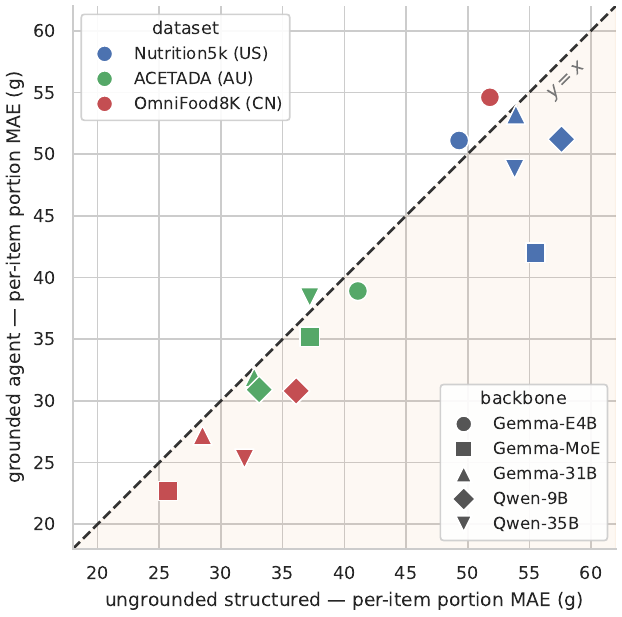}
  \caption{\textbf{Grounding lowers per-item portion error.} Per-item portion MAE, ungrounded structured ($x$) vs.\ grounded agent ($y$), one point per backbone $\times$ dataset; points below $y\!=\!x$ favor grounding (12 of 15).}
  \label{fig:peritem}
\end{figure}

Grounding earns its keep at the item level. The grounded agent attains lower per-item portion error than an ungrounded structured decomposition in 12 of 15 backbone-dataset settings (Figure~\ref{fig:peritem}); per-item energy and macros are mixed, with per-backbone numbers in the supplementary material. With no ground-truth food codes for these benchmarks, per-item nutrient accuracy proxies grounding quality. The decisive difference, though, is not totals, which an ungrounded decomposition matches, but provenance: each item carries the FNDDS code it grounded to, an auditable item-to-code-to-nutrient record a direct estimate cannot give and a dietitian can inspect (Figure~\ref{fig:trace}).

\subsection{Ablations and controls}
\label{sec:ablations}

\begin{table}[htbp]
\centering
\caption{\textbf{Component ablations} (Qwen-35B; N5k and OmniFood; $\Delta$kcal MAE vs.\ the full path, $+$ = worse). Portion is unchanged; calls = MLLM calls/meal (N5k).}
\label{tab:ablations}
\small
\begin{tabular}{l cc c}
\toprule
& \multicolumn{2}{c}{$\Delta$ Energy MAE (kcal)} & \\
\cmidrule(lr){2-3}
Variant & N5k & OmniFood & calls/meal \\
\midrule
\textbf{Open-KNEAD (full)} & $0$ & $0$ & 2.8 \\
\;\;no gate, never pick  & $+26.7$ & $+6.9$ & 1.0 \\
\;\;no gate, always pick & $+1.2$  & $+0.7$ & 3.3 \\
\;\;BM25 only & $+9.9$  & $-1.3$ & 2.9 \\
\;\;dense only & $+6.8$  & $+4.5$ & 2.7 \\
\bottomrule
\end{tabular}
\end{table}

Ablating each component of the decomposition path on Qwen-35B, one of the evaluated backbones from the Qwen family, confirms the design (Table~\ref{tab:ablations}). The selective gate is the operative mechanism: never querying for a code costs 27\,kcal, while always querying recovers nothing beyond the gate yet spends more calls per meal (3.3 vs.\ 2.8). Retrieval does not hinge on one index: dense and lexical each carry a cuisine, so we fuse them. The self-consistency control of Table~\ref{tab:main}, given five calls but none of the agent's structure, does not beat a single direct call, so the gains come from grounded decomposition, not added inference.

\section{Conclusion}
\label{sec:conclusion}

We presented Open-KNEAD, a training-free, knowledge-grounded agentic framework for single-image meal nutrition estimation on open, locally deployable models. Because a modern direct MLLM already matches retrieval pipelines on energy, the contribution is to reclaim grounding's leverage where it still matters: a multi-axis protocol that judges portion and composition rather than energy alone, portion accuracy paired with an auditable item-to-code-to-nutrient record, and a recipe-prior step that restores the invisible cooking-added energy of non-US cuisine.

Across two open MLLM families and three cuisines, the grounded agent improves portion over direct estimation in most settings, surpasses a frontier closed model's direct portion estimate by about $30\%$ on the dietitian-verified set while keeping every image on the local host, recovers the cross-cuisine energy deficit, and outperforms a faithful reimplementation of DietAI24~\cite{dietai24} on every nutrient. The approach has boundaries we characterize in the supplementary material: decompose-and-sum energy trails a holistic estimate off the US-centric knowledge base (handled by routing), the recipe-prior must be cuisine-gated, and the single-image scope forecloses geometric cues. Future work targets remaining headroom: a learned per-meal energy router, an adaptive selective-gate threshold, recipe-prior calibration, broader knowledge-base coverage, and agent-controlled vision tools, inert here but promising with multi-view or scale-referenced capture. We release Open-KNEAD and its agent-ready knowledge base.

\newpage
\clearpage
\appendix
\section{Supplementary Material}
\label{sec:supp}

The main paper reports the full multi-metric tables for a single representative
backbone (Gemma-MoE); the per-ingredient, ablation, and oracle analyses likewise
use one backbone for readability. This supplementary material provides (i) all prompts,
(ii) the complete five-backbone results across every dataset and metric, and
(iii) additional controls.

\subsection{Full dataset landscape}
\label{supp:datasets}
Table~\ref{tab:datasetsfull} expands the main-paper subset to the full landscape we surveyed: only Nutrition5k, ACETADA, and OmniFood8K pair plated meals with measured nutrients, portion/mass, \emph{and} per-ingredient ground truth together, the combination an agentic decomposition system requires for evaluation.
\begin{table*}[t]
\centering
\caption{\textbf{The food-dataset landscape (full).} Evaluating agentic nutrition estimation needs real plated meals with measured nutrients, mass, \emph{and} per-ingredient ground truth; only the three we adopt (bottom) provide all three. ($\sim$ = partial.)}
\label{tab:datasetsfull}
\small
\begin{tabular}{l l c c c c}
\toprule
Dataset & Primary task & \shortstack{Plated\\meals} & \shortstack{Measured\\nutrients} & \shortstack{Portion/\\mass GT} & \shortstack{Per-ingred.\\GT} \\
\midrule
Food-101~\cite{food101}            & classification        & $\times$    & $\times$ & $\times$ & $\times$ \\
Food2K~\cite{food2k}              & classification        & $\times$    & $\times$ & $\times$ & $\times$ \\
UEC-Food256~\cite{uecfood256}         & classification/det.   & $\sim$      & $\times$ & $\times$ & $\times$ \\
VireoFood-172~\cite{vireofood172}       & cls.\ + ingredients   & $\times$    & $\times$ & $\times$ & $\sim$ \\
FoodSeg103~\cite{foodseg103}          & segmentation          & \checkmark  & $\times$ & $\times$ & $\sim$ \\
UNIMIB2016~\cite{unimib2016}          & tray segmentation     & \checkmark  & $\times$ & $\times$ & $\times$ \\
Recipe1M+~\cite{recipe1m}           & recipe retrieval      & $\times$    & $\sim$   & $\times$ & $\sim$ \\
ECUSTFD~\cite{ecustfd}             & volume (fiducial)     & $\times$    & $\times$ & \checkmark & $\times$ \\
FastFood~\cite{qi2025vif2} & cls.\ + nutrition & $\sim$  & $\sim$   & $\times$ & $\sim$ \\
FoodNExTDB~\cite{romerotapiador2025ready} & expert categories & \checkmark & $\times$ & $\times$ & $\sim$ \\
MetaFood3D~\cite{metafood3d} & 3D single items & $\times$ & \checkmark & \checkmark & $\sim$ \\
SNAPMe~\cite{snapme} & nutrition (user-annotated) & \checkmark & $\sim$ & $\sim$ & $\sim$ \\
\midrule
\textbf{Nutrition5k}~\cite{thames2021nutrition5k} & nutrition (weighed) & \checkmark & \checkmark & \checkmark & \checkmark \\
\textbf{ACETADA}~\cite{acetada} & nutrition (dietitian) & \checkmark & \checkmark & \checkmark & \checkmark \\
\textbf{OmniFood8K}~\cite{yu2026omnifood8k} & nutrition (weighed) & \checkmark & \checkmark & \checkmark & \checkmark \\
\bottomrule
\end{tabular}
\end{table*}

\subsection{Prompts}
\label{supp:prompts}
We give the verbatim prompts for the four model calls Open-KNEAD makes: the Path~A direct estimate, the visible-item decomposition (Path~B), the recipe-prior, and the selective candidate pick. Tokens in braces (e.g., \texttt{\{visible\}}, \texttt{\{item\_name\}}, \texttt{\{candidate\_block\}}) are filled at inference; angle-bracketed values denote the requested JSON fields.

\noindent\textbf{Path A: direct estimate.}
\begin{lstlisting}[style=prompt]
You are a nutrition estimation assistant.

Look at the meal image and estimate the total nutrition for the meal as served.
Return ONLY a JSON object with these keys and numeric values:

{
  "kcal": <total kilocalories, float>,
  "protein_g": <total protein in grams, float>,
  "fat_g": <total fat in grams, float>,
  "carb_g": <total carbohydrates in grams, float>,
  "portion_g": <estimated total served mass in grams, float or null>,
  "confidence": <float between 0 and 1>,
  "rationale": "<one short sentence>"
}

Do not include any text outside the JSON object.
\end{lstlisting}

\noindent\textbf{Path B: visible-item decomposition.}
\begin{lstlisting}[style=prompt]
You are a meal-image food parser.

Look at the meal image and list each visible food item.
For each item, give a short retrieval-friendly food description (no nutrient
numbers) and an estimated portion mass in grams.

Return ONLY this JSON object:

{
  "items": [
    {
      "name": "<short food name, e.g. 'grilled chicken breast'>",
      "query": "<retrieval query text>",
      "portion_estimate_g": <float>,
      "confidence": <float between 0 and 1>
    }
  ]
}

Do not include any text outside the JSON object.
\end{lstlisting}

\noindent\textbf{Recipe-prior.}
\begin{lstlisting}[style=prompt]
You are a nutrition-aware cooking expert.

A meal contains these VISIBLE food items:
{visible}

List the typical COOKING-ADDED ingredients a meal like this usually contains but
that are NOT visually obvious in a photo -- cooking oil, butter, ghee, lard,
sauces, dressings, syrups, or added sugar. These are the hidden, calorie-dense
components a person cannot see.

Rules:
- Only add ingredients NOT already implied by the visible list.
- Give an approximate amount in grams, appropriate for the meal's size.
- If the meal is just raw / whole / uncooked foods (fresh fruit, a plain salad
  with no dressing, plain bread, raw vegetables), return an EMPTY list -- do NOT
  invent ingredients.

Return ONLY this JSON:
{"items": [{"name": "<ingredient>", "query": "<retrieval query>", "portion_estimate_g": <float>}]}
\end{lstlisting}

\noindent\textbf{Selective candidate pick.}
\begin{lstlisting}[style=prompt]
You are choosing FNDDS food code(s) that best match an item visible in a meal image.

Item name: {item_name}
Estimated portion: {portion_g} g

Candidate FNDDS entries (numbered):
{candidate_block}

Instructions:
- Choose the smallest set of candidate numbers (1-based) whose descriptions match the food actually visible.
- Prefer one candidate when possible. Return up to {max_picks}.
- If none of the candidates plausibly match the visible food, return an empty list.

Return ONLY this JSON object:

{
  "picks": [<candidate_number>, ...],
  "reason": "<one short sentence>"
}

Do not include any text outside the JSON object.
\end{lstlisting}

\subsection{DietAI24 reimplementation}
\label{supp:dietai24}
The original DietAI24~\cite{dietai24} pipeline targets GPT-4V, which is no longer served by the OpenAI API. We reimplement its retrieve-then-pick procedure on open-weight backbones, preserving its structure (per-item retrieval from a nutrition database, then model selection of a matching code) while substituting our FNDDS knowledge base and BM25/BGE-M3 retrieval, so every method is compared on the same models and knowledge base. Paired on shared dishes, our grounding comparator improves on this reimplementation across every backbone and dataset, on energy and portion alike (Table~\ref{tab:undergrad}), confirming the gains generalize beyond the featured backbone.

\begin{table}[t]
\centering
\caption{\textbf{Open-KNEAD's grounding vs.\ the prior pipeline, paired on shared dishes.} An independent faithful reimplementation of DietAI24~\cite{dietai24} vs.\ our same-knowledge-base grounding comparator (RAG\,$+$\,VLM pick), energy and portion MAE, per backbone. \textbf{Bold} = lower; all reported differences have 95\% bootstrap CIs excluding zero. OmniFood is available only for Gemma-MoE in the reimplementation.}
\label{tab:undergrad}
\small
\setlength{\tabcolsep}{5pt}
\begin{tabular}{ll cc cc}
\toprule
& & \multicolumn{2}{c}{energy MAE (kcal)\,$\downarrow$} & \multicolumn{2}{c}{portion MAE (g)\,$\downarrow$} \\
\cmidrule(lr){3-4}\cmidrule(lr){5-6}
Backbone & Dataset & DietAI24 & ours & DietAI24 & ours \\
\midrule
Gemma-MoE & N5k      & 154.8 & \textbf{103.3} & 123.4 & \textbf{71.8} \\
          & ACETADA  & 249.2 & \textbf{182.8} & 367.5 & \textbf{163.3} \\
          & OmniFood & 224.6 & \textbf{186.6} & 102.0 & \textbf{41.0} \\
\midrule
Gemma-E4B & N5k      & 266.2 & \textbf{200.4} & 219.9 & \textbf{143.3} \\
          & ACETADA  & 322.7 & \textbf{225.9} & 465.4 & \textbf{235.3} \\
\midrule
Qwen-9B   & N5k      & 346.9 & \textbf{103.2} & 259.1 & \textbf{81.1} \\
          & ACETADA  & 284.8 & \textbf{182.7} & 350.5 & \textbf{238.4} \\
\midrule
Qwen-35B  & N5k      & 191.5 & \textbf{97.7}  & 155.7 & \textbf{80.4} \\
          & ACETADA  & 260.5 & \textbf{170.3} & 435.9 & \textbf{208.6} \\
\bottomrule
\end{tabular}
\end{table}

\subsection{Full main results (all backbones)}
\label{supp:fullmain}
Table~\ref{tab:mainfull} reports energy, severe-error rate, and portion for the direct estimate and the grounded agent on all five backbones. The portion gain of the main paper holds throughout: the agent lowers portion MAE in $12$ of $15$ backbone-dataset settings, the exceptions being near-ties on the smallest backbone (Gemma-E4B). Energy here is the raw decompose-and-sum, before the cuisine routing of \S\ref{supp:recipefull}.
\begin{table*}[t]
\centering
\caption{\textbf{Per-backbone results: direct vs.\ grounded agent} (energy/portion MAE; sev = \% of meals off by $>$100\,kcal; full splits). \textbf{Bold} = agent better than direct. Energy here is the raw decompose-and-sum; the deployed system routes energy by cuisine (Table~\ref{tab:recipefull}) and takes portion from the agent.}
\label{tab:mainfull}
\small
\setlength{\tabcolsep}{4pt}
\begin{tabular}{l cc cc cc}
\toprule
& \multicolumn{2}{c}{kcal MAE\,$\downarrow$} & \multicolumn{2}{c}{sev (\%)\,$\downarrow$} & \multicolumn{2}{c}{portion MAE (g)\,$\downarrow$} \\
\cmidrule(lr){2-3}\cmidrule(lr){4-5}\cmidrule(lr){6-7}
Backbone & direct & agent & direct & agent & direct & agent \\
\midrule
\multicolumn{7}{l}{\emph{Nutrition5k (US)}} \\
Gemma-E4B & 134.8 & 195.7 & 45.6 & 55.2 & 121.1 & 123.0 \\
Gemma-MoE & 121.9 & \textbf{109.7} & 38.8 & \textbf{28.7} & 80.2 & \textbf{66.7} \\
Gemma-31B & 107.3 & 115.9 & 31.3 & 35.2 & 95.7 & \textbf{69.4} \\
Qwen-9B   & 116.7 & 117.8 & 36.0 & \textbf{32.6} & 92.0 & \textbf{74.9} \\
Qwen-35B  & 113.0 & \textbf{106.9} & 30.8 & \textbf{28.7} & 81.7 & \textbf{72.9} \\
\midrule
\multicolumn{7}{l}{\emph{ACETADA (AU)}} \\
Gemma-E4B & 236.2 & 247.7 & 63.9 & 68.4 & 313.7 & \textbf{244.7} \\
Gemma-MoE & 163.5 & 193.3 & 59.6 & 61.6 & 254.1 & \textbf{167.3} \\
Gemma-31B & 171.8 & 199.4 & 63.8 & 65.6 & 150.5 & \textbf{147.5} \\
Qwen-9B   & 159.6 & 195.8 & 52.7 & 64.8 & 331.7 & \textbf{245.4} \\
Qwen-35B  & 167.1 & 177.4 & 54.1 & 58.0 & 301.6 & \textbf{222.5} \\
\midrule
\multicolumn{7}{l}{\emph{OmniFood8K (CN)}} \\
Gemma-E4B & 122.4 & 162.4 & 47.3 & 58.1 & 119.8 & 121.1 \\
Gemma-MoE & 155.2 & 168.6 & 61.2 & 63.1 & 62.4 & \textbf{43.5} \\
Gemma-31B & 168.7 & \textbf{159.5} & 63.1 & \textbf{61.2} & 95.6 & \textbf{60.8} \\
Qwen-9B   & 129.2 & 151.9 & 47.7 & 57.3 & 112.2 & \textbf{68.6} \\
Qwen-35B  & 139.2 & 149.8 & 51.8 & 57.2 & 97.0 & \textbf{57.0} \\
\bottomrule
\end{tabular}
\end{table*}

\subsection{Portion MAE, all backbones}
\label{supp:portion}
Table~\ref{tab:portion} isolates portion MAE with paired significance: bold marks the settings where the grounded agent beats direct with a $95\%$ bootstrap CI excluding zero. The privacy result of the main paper appears here as well---on ACETADA the local Gemma-31B agent ($147$\,g) beats both the gpt-5.5 ($212$\,g) and Gemini ($311$\,g) direct estimates.
\begin{table}[t]
\centering
\caption{\textbf{Portion MAE (g): grounded agent vs.\ direct, all backbones.} \textbf{Bold} = significantly better (paired 95\% CI excludes 0). \emph{Privacy:} on ACETADA the local Gemma-31B agent (147) beats gpt-5.5 (212; $-$30\%) and Gemini (311; $-$53\%) direct estimates.}
\label{tab:portion}
\small
\setlength{\tabcolsep}{4.5pt}
\begin{tabular}{l cc cc cc}
\toprule
Portion MAE (g)\,$\downarrow$ & \multicolumn{2}{c}{N5k (US)} & \multicolumn{2}{c}{ACETADA (AU)} & \multicolumn{2}{c}{OmniFood (CN)} \\
\cmidrule(lr){2-3}\cmidrule(lr){4-5}\cmidrule(lr){6-7}
Backbone & direct & agent & direct & agent & direct & agent \\
\midrule
Gemma-E4B  & 121.1 & 123.0 & 313.7 & \textbf{244.7} & 119.8 & 121.1 \\
Gemma-MoE  & 80.2 & \textbf{66.7} & 254.1 & \textbf{167.3} & 62.4 & \textbf{43.5} \\
Gemma-31B  & 95.7 & \textbf{69.4} & 150.5 & 147.5 & 95.6 & \textbf{60.8} \\
Qwen-9B    & 92.0 & \textbf{74.9} & 331.7 & \textbf{245.4} & 112.2 & \textbf{68.6} \\
Qwen-35B   & 81.7 & \textbf{72.9} & 301.6 & \textbf{222.5} & 97.0 & \textbf{57.0} \\
\bottomrule
\end{tabular}
\end{table}

\subsection{Recipe-prior signed bias (all backbones)}
\label{supp:recipefull}
Table~\ref{tab:recipefull} gives the signed energy bias (mean $\hat{e}-e$) for the direct estimate, the visible-only agent, and the agent with the recipe-prior. The prior corrects the negative bias (underestimation) on OmniFood, where invisible cooking energy is present, and overshoots on Nutrition5k and ACETADA, where it is not---which is why the deployed system gates it by cuisine.
\begin{table}[t]
\centering
\caption{\textbf{Recipe-prior signed energy bias} (kcal; all backbones; full splits). The metric is signed (mean $\hat{e}-e$): its sign shows over- vs.\ under-estimate and $0$ is unbiased, so there is no ``lower is better'' direction. All changes: 95\% CI excludes 0.}
\label{tab:recipefull}
\small
\begin{tabular}{l ccc}
\toprule
Backbone & direct & agent & $+$recipe \\
\midrule
\multicolumn{4}{l}{\emph{OmniFood8K (CN) --- deficit present}} \\
Gemma-E4B & $-56.0$  & $+16.2$  & $+116.7$ \\
Gemma-MoE & $-85.5$  & $-123.2$ & $\mathbf{+37.3}$ \\
Gemma-31B & $-131.5$ & $-98.2$  & $+93.6$ \\
Qwen-9B   & $-86.1$  & $-77.5$  & $\mathbf{-14.2}$ \\
Qwen-35B  & $-110.2$ & $-113.7$ & $\mathbf{-21.7}$ \\
\midrule
\multicolumn{4}{l}{\emph{Nutrition5k (US) --- no deficit (overshoots)}} \\
Gemma-E4B & $+44.6$  & $+120.8$ & $+206.0$ \\
Gemma-MoE & $+22.5$  & $-10.8$  & $+137.9$ \\
Gemma-31B & $-0.4$   & $+3.3$   & $+133.0$ \\
Qwen-9B   & $+16.0$  & $+4.1$   & $+51.9$ \\
Qwen-35B  & $-19.2$  & $-7.4$   & $+50.6$ \\
\midrule
\multicolumn{4}{l}{\emph{ACETADA (AU) --- no deficit (overshoots)}} \\
Gemma-E4B & $-200.5$ & $+74.4$  & $+219.8$ \\
Gemma-MoE & $+21.0$  & $+15.0$  & $+253.1$ \\
Gemma-31B & $+64.0$  & $+52.1$  & $+235.4$ \\
Qwen-9B   & $-8.4$   & $+41.9$  & $+159.7$ \\
Qwen-35B  & $-100.0$ & $-53.5$  & $+33.7$ \\
\bottomrule
\end{tabular}
\end{table}

\subsection{Per-ingredient grounding (all backbones)}
\label{supp:peritemfull}
Tables~\ref{tab:peritemfull} and~\ref{tab:peritem} compare per-ingredient MAE for the grounded agent against the ungrounded structured decomposition on ground-truth-matched items. Grounding lowers per-item portion error in $12$ of $15$ settings; per-item energy and macronutrients are mixed, mirroring the meal-level totals.
\begin{table}[t]
\centering
\caption{\textbf{Per-ingredient MAE, all backbones: grounded agent vs.\ ungrounded structured} (GT-matched items). \textbf{Bold} = grounded lower.}
\label{tab:peritemfull}
\small
\begin{tabular}{l cc cc}
\toprule
& \multicolumn{2}{c}{portion MAE (g)\,$\downarrow$} & \multicolumn{2}{c}{kcal MAE\,$\downarrow$} \\
\cmidrule(lr){2-3}\cmidrule(lr){4-5}
Backbone & agent & struct. & agent & struct. \\
\midrule
\multicolumn{5}{l}{\emph{Nutrition5k}} \\
Gemma-E4B & 51.1 & \textbf{49.3} & 74.2 & \textbf{55.2} \\
Gemma-MoE & \textbf{42.0} & 55.5 & \textbf{52.5} & 66.5 \\
Gemma-31B & \textbf{53.2} & 53.9 & 67.2 & \textbf{65.3} \\
Qwen-9B   & \textbf{51.2} & 57.6 & 68.6 & \textbf{64.0} \\
Qwen-35B  & \textbf{48.8} & 53.8 & \textbf{62.9} & 67.6 \\
\midrule
\multicolumn{5}{l}{\emph{ACETADA}} \\
Gemma-E4B & \textbf{38.9} & 41.1 & 57.3 & \textbf{41.1} \\
Gemma-MoE & \textbf{35.2} & 37.2 & 41.2 & 41.2 \\
Gemma-31B & \textbf{31.9} & 32.7 & 40.2 & \textbf{38.9} \\
Qwen-9B   & \textbf{30.9} & 33.1 & 52.5 & \textbf{36.3} \\
Qwen-35B  & 38.4 & \textbf{37.2} & 36.8 & \textbf{35.7} \\
\midrule
\multicolumn{5}{l}{\emph{OmniFood8K}} \\
Gemma-E4B & 54.6 & \textbf{51.8} & \textbf{94.5} & 98.7 \\
Gemma-MoE & \textbf{22.7} & 25.7 & 64.5 & \textbf{59.9} \\
Gemma-31B & \textbf{27.2} & 28.5 & 63.4 & \textbf{63.0} \\
Qwen-9B   & \textbf{30.8} & 36.1 & \textbf{61.6} & 77.0 \\
Qwen-35B  & \textbf{25.3} & 31.9 & \textbf{52.3} & 56.0 \\
\bottomrule
\end{tabular}
\end{table}

\begin{table}[t]
\centering
\caption{\textbf{Per-ingredient MAE: grounded agent vs.\ ungrounded structured} (Gemma-MoE, GT-matched items). \textbf{Bold} = lower (better).}
\label{tab:peritem}
\small
\setlength{\tabcolsep}{4pt}
\begin{tabular}{l cc cc cc}
\toprule
& \multicolumn{2}{c}{\textbf{N5k}} & \multicolumn{2}{c}{\textbf{ACETADA}} & \multicolumn{2}{c}{\textbf{OmniFood}} \\
\cmidrule(lr){2-3}\cmidrule(lr){4-5}\cmidrule(lr){6-7}
Per-item MAE\,$\downarrow$ & agent & struct. & agent & struct. & agent & struct. \\
\midrule
portion (g) & \textbf{42.0} & 55.5 & \textbf{35.2} & 37.2 & \textbf{22.7} & 25.7 \\
kcal        & \textbf{52.5} & 66.5 & 41.2          & 41.2 & 64.5          & \textbf{59.9} \\
protein (g) & 2.6           & \textbf{2.5} & 1.8 & \textbf{1.3} & \textbf{3.3} & 3.5 \\
fat (g)     & \textbf{3.4}  & 4.5  & \textbf{1.9}  & 2.0 & 4.1 & \textbf{3.4} \\
carb (g)    & \textbf{4.4}  & 5.2  & 6.0           & \textbf{5.9} & \textbf{6.4} & 6.7 \\
\bottomrule
\end{tabular}
\end{table}

\subsection{Meal-level macronutrients}
\label{supp:macros}
Table~\ref{tab:macros} reports protein, fat, and carbohydrate MAE and pMAE for every method on the featured Gemma-MoE backbone. Because macronutrients route together with energy, Open-KNEAD tracks the direct estimate except where the recipe-prior fires, most visibly the OmniFood fat that recovers invisible cooking oil.
\begin{table*}[t]
\centering
\caption{\textbf{Meal-level macronutrient estimation} (Gemma-MoE; full test splits). MAE and pMAE for protein, fat, and carbohydrate. \textbf{Bold} = best, \underline{underline} = second best per dataset (lower better). Macronutrients are routed together with energy (Path~A, or Path~B$+$recipe by cuisine), so Open-KNEAD tracks the direct estimate except where the recipe-prior applies, most visibly OmniFood fat (oil).}
\label{tab:macros}
\small
\setlength{\tabcolsep}{6pt}
\begin{tabular}{l cc cc cc}
\toprule
& \multicolumn{2}{c}{\textbf{Protein}} & \multicolumn{2}{c}{\textbf{Fat}} & \multicolumn{2}{c}{\textbf{Carbohydrate}} \\
\cmidrule(lr){2-3}\cmidrule(lr){4-5}\cmidrule(lr){6-7}
Method & MAE (g)\,$\downarrow$ & pMAE\%\,$\downarrow$ & MAE (g)\,$\downarrow$ & pMAE\%\,$\downarrow$ & MAE (g)\,$\downarrow$ & pMAE\%\,$\downarrow$ \\
\midrule
\multicolumn{7}{l}{\emph{Nutrition5k (US)}} \\
Direct prompting                 & 7.1 & 44.5 & 9.0 & 65.7 & 11.7 & 61.6 \\
Self-consistency direct          & 7.1 & 44.8 & 8.9 & 65.3 & 11.7 & 61.1 \\
Structured decomposition (no KB) & \textbf{6.5} & \textbf{40.7} & 8.8 & 64.4 & \textbf{9.6} & \textbf{50.5} \\
RAG, weighted top-$K$            & 7.2 & 45.3 & 9.2 & 67.0 & 12.2 & 64.0 \\
RAG\,$+$\,VLM pick               & \underline{7.0} & \underline{43.6} & \underline{8.7} & \underline{63.3} & 9.8 & 51.2 \\
\textbf{Open-KNEAD (ours)}       & \underline{7.0} & 44.0 & \textbf{8.5} & \textbf{62.2} & \underline{9.7} & \underline{50.8} \\
\midrule
\multicolumn{7}{l}{\emph{ACETADA (AU)}} \\
Direct prompting                 & 12.7 & 36.3 & 9.1 & 37.8 & \underline{21.2} & \underline{27.8} \\
Self-consistency direct          & 12.7 & 36.3 & \underline{9.0} & \underline{37.6} & \textbf{20.7} & \textbf{27.3} \\
Structured decomposition (no KB) & \underline{12.2} & \underline{35.1} & \underline{9.0} & 37.7 & 24.1 & 31.7 \\
RAG, weighted top-$K$            & \textbf{12.1} & \textbf{34.7} & 13.1 & 54.6 & 30.5 & 40.1 \\
RAG\,$+$\,VLM pick               & 13.7 & 39.3 & \textbf{8.9} & \textbf{37.2} & 24.5 & 32.2 \\
\textbf{Open-KNEAD (ours)}       & 12.7 & 36.3 & 9.1 & 37.8 & \underline{21.2} & \underline{27.8} \\
\midrule
\multicolumn{7}{l}{\emph{OmniFood8K (CN)}} \\
Direct prompting                 & 7.1 & 58.2 & \underline{16.6} & 60.5 & 19.3 & 84.3 \\
Self-consistency direct          & 7.1 & 58.0 & \underline{16.6} & \underline{60.4} & 19.3 & 84.2 \\
Structured decomposition (no KB) & 6.6 & 53.6 & 16.8 & 61.3 & \underline{14.7} & \underline{64.4} \\
RAG, weighted top-$K$            & \textbf{6.4} & \textbf{52.2} & 18.0 & 65.6 & 17.5 & 76.4 \\
RAG\,$+$\,VLM pick               & \underline{6.5} & \underline{53.0} & 18.1 & 65.9 & \textbf{13.7} & \textbf{59.9} \\
\textbf{Open-KNEAD (ours)}       & 6.7 & 55.0 & \textbf{10.8} & \textbf{39.6} & 17.9 & 78.4 \\
\bottomrule
\end{tabular}
\end{table*}

\subsection{Limitations}
\label{sec:boundaries}

Open-KNEAD operates from a single image by design and is, to our knowledge, the first agentic framework of this kind. Within this scope we probed vision mechanisms that might sharpen estimates and found them inert (Table~\ref{tab:negatives}): monocular depth~\cite{piccinelli2024unidepth} and projected food area do not recover mass (area-to-mass correlation $-0.02$), promptable segmentation~\cite{kirillov2023sam, sam3} isolates items but not their mass, crop-to-code matching in a shared vision-language space (SigLIP\,2~\cite{tschannen2025siglip2} embeddings of segmented crops, with set-of-marks~\cite{yang2023som}) collapses under a shuffle control (MRR $0.949$ vs.\ $0.962$), and extra camera views do not help a model that cannot fuse them geometrically. We therefore estimate portion from the visible decomposition and leave richer-capture vision tools, which need multi-view or a scale reference~\cite{vinod2024scaling,chen2026implicit}, to future work.

\begin{table}[t]
\centering
\caption{\textbf{Vision mechanisms that did not improve estimation} (single-image, local-inference scope). Each was probed and treated as a boundary, not adopted.}
\label{tab:negatives}
\small
\begin{tabular}{p{0.27\linewidth} p{0.64\linewidth}}
\toprule
Approach & Outcome \\
\midrule
Monocular metric depth~\cite{piccinelli2024unidepth} ($\to$ volume/mass) & Depth scale unreliable at plate scale ($\sim$1.2\,m estimated vs.\ $\sim$0.35\,m true; scene noise $\gg$ food height); projected-area-to-mass correlation $-0.02$. \\
Multi-view fusion (extra camera views) & Worsens portion for a model that cannot fuse views geometrically: $+13.6$\,g (N5k), $+27.2$\,g (OmniFood) MAE vs.\ single view. \\
Promptable segmentation~\cite{kirillov2023sam,sam3} & Isolates items cleanly but does not address mass, the bottleneck. \\
Visual code disambiguation (SigLIP\,2~\cite{tschannen2025siglip2} crop\,$\leftrightarrow$\,FNDDS, incl.\ set-of-marks~\cite{yang2023som}) & Zero-shot cross-modal cosine is at chance (MRR 0.31 vs.\ 0.29 random, 0.55 text); a learned head's gain vanishes under a crop-shuffle control (MRR 0.949 vs.\ 0.962), i.e.\ crop-independent. \\
\bottomrule
\end{tabular}
\end{table}

The deployed system has boundaries too. Decompose-and-sum energy carries a density bias away from the US-centric knowledge base: on the dietitian-verified Australian dataset the summed total trails the holistic estimate even under oracle labels and portions ($\approx 201$ vs.\ $152$\,kcal MAE), so the deployed routing sources that energy from the direct path. This routing is cuisine-level, not per-meal, because a leakage-safe per-meal gate (path agreement or per-item grounding ambiguity) did not separate the meals where decompose-sum wins: the unreliability is a distribution-level density bias, not the per-meal candidate disagreement such gates measure. The recipe-prior likewise adds a roughly constant invisible-energy budget, so it is gated on recognized cuisine rather than applied to every meal. The single-image input is a deliberate choice to minimize logging burden that forecloses richer geometric cues, and we report medians alongside means because the benchmarks contain physically inconsistent labels; broader knowledge-base coverage of non-US foods we leave to future work.

\subsection{Failure modes}
\label{supp:failures}
Beyond these system-level boundaries, the agent shows a few recurring per-meal failures. First, mis-recognition of visually ambiguous items: a decomposed name can match a plausible but wrong food, for example an aromatic read as a prepared product (onion grounded to an onion dip, or salt to pistachio); the meal total can still land near the target when such errors offset, which is itself fragile. Second, fine-grained code selection: when candidates are nutritionally close the gate commits the weighted mean, which is correct on average but wrong for an item whose true code sits in the tail of the candidate set. Third, portion errors on occluded, layered, or mixed dishes, where per-item mass is hard to read from a single view. Fourth, recipe-prior overshoot where no invisible additions are present, which the cuisine-level gating is designed to avoid.


\newpage
\clearpage
{
    \small
    \bibliographystyle{ieeenat_fullname}
    \bibliography{main}
}

\end{document}